\documentclass[11pt]{article}

\usepackage[margin=1in]{geometry}
\usepackage{amsmath,amssymb,amsfonts}
\usepackage{graphicx}
\usepackage{booktabs}
\usepackage{multirow}
\usepackage{microtype}
\usepackage{xcolor}
\usepackage{hyperref}
\usepackage{cleveref}
\usepackage{natbib}

\title{\textbf{Seasonality-Aware Hybrid Convolutional Transformer for Antarctic Sea Ice Concentration Forecasting}}

\author{
Danyang Li$^{1}$ \quad
John Taylor$^{1,*}$ \quad
Thang Bui$^{1}$ \quad
Quanling Deng$^{1,2,*}$
\\[6pt]
$^{1}$School of Computing, Australian National University,
Canberra, ACT 2601, Australia
\\
$^{2}$Yau Mathematical Sciences Center, Tsinghua University,
Beijing, China
\\[4pt]
\texttt{
danyang.li@anu.edu.au,
john.taylor@anu.edu.au,} \\ \texttt{
thang.bui@anu.edu.au,
qldeng@tsinghua.edu.cn, quanling.deng@anu.edu.au
}
\\[3pt]
$^{*}$Corresponding authors
}

\date{}
\begin{document}
\maketitle
\begin{abstract}
  Antarctic sea ice concentration (SIC) forecasting is an important yet challenging task due to the coexistence of complex spatial structure, long-range temporal dependencies, and strong seasonal variability. Conventional convolution-based models are effective at capturing local spatial patterns, but often have limited ability to model long-term temporal evolution. To address these challenges, we build on a hybrid convolutional transformer forecasting framework for monthly Antarctic SIC forecasting. This framework combines convolutional encoding for spatial feature extraction with space-time factorised self-attention for SIC modelling. We further introduce two season-based mechanisms: a month-aware positional encoding that injects calendar-month information into the token representation, and a seasonal temporal bias that encourages attention to periodically related historical states. Experimental results show that the proposed framework achieves better performance than convolutional and recurrent baselines as well as ECMWF's physics-based dynamical model SEAS5 under both classification and regression metrics. Ablation studies further indicate that the seasonality-aware components provide consistent additional gains in both short- and long-horizon prediction. These results demonstrate the value of combining convolutional structures, attention mechanisms, and periodic prior information for Antarctic SIC forecasting.
\end{abstract}

\section{Introduction}
\label{sec:intro}
Antarctic sea ice is a critical component of the Earth’s climate system, influencing the surface energy balance and ocean-atmosphere interactions \cite{turner2015recent, parkinson201940}. Unlike the Arctic, Antarctic sea ice exhibits pronounced interannual variability and strong regional heterogeneity, with record lows and partial recoveries observed in recent decades \cite{purich2023record, josey2024record}. Reliable prediction of Antarctic sea ice concentration (SIC) on seasonal timescales is therefore important for climate monitoring and downstream applications such as polar logistics and marine operations. Yet Antarctic SIC forecasting remains challenging due to complex atmosphere-ocean-sea-ice interactions, sharp gradients at the ice edge, and a strong but non-stationary seasonal cycle \cite{hobbs2016review}, making it a difficult spatio-temporal sequence-to-sequence learning problem.

Sea ice forecasting has traditionally relied on coupled numerical models that explicitly simulate thermodynamic and dynamic processes, including ocean-ice interactions, wind forcing, and heat exchange \cite{numerical-mellor, numerical-blanchard, numerical-wayand}. Representative systems include Global Sea Ice 6.0, a CICE-based sea-ice model configuration developed for the Met Office Global Coupled model \cite{rae2015development}. Seasonal Forecasting System 5 (SEAS5) developed by the European Centre for Medium-Range Weather Forecasts (ECMWF), which provides operational seasonal forecasts \cite{johnson2019seas5}. While these approaches provide physically consistent predictions, they are computationally expensive and often require substantial effort for calibration and data assimilation, which limit their flexibility for rapid or high-frequency forecasting \cite{liu2021extended}.

As computationally efficient alternatives, statistical and empirical methods have also been explored for sea-ice prediction based on historical observations and climate predictors, such as multiple linear regression, linear empirical models, Bayesian regression, and linear Markov models \cite{drobot2006long, horvath2020bayesian, yuan2016arctic, lindsay2008seasonal}.  However, many traditional statistical models rely on simplified linear or probabilistic assumptions, which may limit their ability to represent complex nonlinear interactions within the sea-ice system, particularly under extreme or rapidly changing conditions \cite{zhu2023deep}.

Recent advances in machine learning have introduced data-driven alternatives for sea ice prediction. Convolutional neural networks (CNNs) and their variants have demonstrated skill by learning spatial features directly from gridded SIC fields \cite{ren2022data, icener-ori, liu2021extended}. Notably, IceNet \cite{icener-ori} showed that UNet-based architectures can leverage multi-source climate inputs to achieve competitive seasonal forecasting performance. However, such 2D CNN-based approaches primarily operate as spatial pattern recognisers and lack explicit mechanisms to model long-range temporal dependencies, which are critical for multi-month prediction. To incorporate temporal dynamics, other models with spatio-temporal recurrent architectures such as Convolutional Long Short-Term Memory (ConvLSTM) and Predictive Recurrent Neural Networks (PredRNN) introduce temporal memory \cite{shi2015convolutional,taylor2022deep,lin2020self,wang2022predrnn}. By embedding convolutional operations within recurrent units, these models preserve spatial structure while evolving temporal states. Nevertheless, their reliance on local convolutional kernels may limit their ability to capture spatial interactions across the polar domain, and training can become increasingly challenging for longer forecasting horizons \cite{zhu2023deep, liu2021extended}.

More recently, attention-based models and Transformers have emerged as a promising alternative for modelling long-range dependencies in spatio-temporal systems \cite{vaswani2017attention,dosovitskiy2021an}. Self-attention mechanisms enable global information exchange across both space and time, which is particularly advantageous for capturing large-scale spatio-temporal dependencies in climate fields  \cite{gao2022earthformer, nguyen2023climax}. However, directly applying global self-attention to high-resolution geophysical fields remains computationally challenging due to the quadratic complexity of self-attention with respect to the number of spatial tokens. Notably, works such as TimeSformer \cite{bertasius2021space} demonstrated that factorising global self-attention into separate spatial and temporal operations drastically reduces computational complexity while maintaining state-of-the-art performance. Similar factorised and local spatio-temporal attention designs have also been explored in video vision Transformers \cite{arnab2021vivit, liu2022video}. However, these generic video architectures are not specifically designed for geophysical forecasting and do not explicitly encode domain-specific priors such as seasonal cycles, motivating further adaptations to both manage the spatial token scale and incorporate seasonal information.

To address these challenges, we propose a convolutional transformer framework for Antarctic SIC forecasting that combines convolutional spatial feature extraction with Transformer-based spatio-temporal dependency modelling. Our main contributions are as following:
\begin{enumerate}
\item Convolutional transformer architecture with bottleneck attention: We develop a convolutional encoder-decoder for high-resolution spatial reconstruction, and adopt a Transformer bottleneck that performs global spatio-temporal mixing only at a compressed resolution, enabling multi-month forecasting at feasible GPU cost.

\item Month-aware Positional Encoding (MonthPE): To avoid an large 3D positional lookup table, we factorise positional and seasonal information into three distinct embeddings. This includes a standard spatial encoding, standard temporal encoding, and a Fourier-based seasonal encoding that encodes the strong seasonal periodicity of monthly sea ice.

\item Annually Temporal Attention Bias (SeasonBias): We design a divided space-time attention scheme to manage computational complexity. We augment the temporal attention logits with a learnable seasonal bias based on the circular distance of the annual cycle, directly modulating query-key interactions into seasonal states.
\end{enumerate}

\section{Problem Statement and Data}
\label{sec:prob}
We frame Antarctic SIC forecasting as a spatio-temporal sequence-to-sequence prediction problem, where the model produces forecasts of future monthly SIC fields. Let the polar environment at a given time $t$ be represented by a 2D spatial grid $X_t \in \mathbb{R}^{H \times W}$ denoting the monthly SIC input field. $H$ and $W$ denote the spatial height and width.
Given a historical observation window of $N_{in}$ consecutive time steps, $\mathcal{X} =\{X_{t-(N_{in}-1)\Delta t}, X_{t-(N_{in}-2)\Delta t},\dots, X_{t}\} \in \mathbb{R}^{N_{in} \times H \times W}$, where $\Delta t$ is time step size and in default, we set it as one month.
The objective is to learn a mapping function $\mathcal{F}_\theta(\cdot): \mathbb{R}^{ N_{in}\times H\times W} \rightarrow\mathbb{R}^{N_{out}\times H\times W}$, parameterised by a transformer-based neural network $\mathcal{T}_{\theta}(\cdot): \mathbb{R}^{ N_{in}\times H\times W} \rightarrow\mathbb{R}^{H\times W}$ (see Section \ref{sec:fram}) with autoregression, to predict the future states over the next $N_{out}$ forecast time steps: 
\begin{equation}\label{modeleq}
    \hat{\mathcal{X}} = \{\hat{X}_{t+\Delta t}, \hat{X}_{t+2\Delta t} , \dots, \hat{X}_{t+N_{out}\Delta t}\} = \mathcal{F}_\theta(\mathcal{X}),
\end{equation}
where $\theta$ characterises the neural network.

To train and evaluate our framework, we utilise the Sea Ice Index, Version 3 dataset provided by the National Snow and Ice Data Center (NSIDC) \cite{seaiceindex}. This dataset is derived from a continuous, multi-decadal satellite record of passive microwave observations (including SMMR, SSM/I, and SSMIS sensors), providing a highly consistent benchmark for climatological machine learning. The dataset provides monthly observations mapped to a polar stereographic projection at a spatial resolution of $25 \times 25 \text{km}^2$ per pixel in 2D; see examples in Figure \ref{fig:structure}. For our network architecture, all Antarctic gridded data are uniformly resized to a fixed spatial resolution of size $H=352$ and $W=352$. We use data from 1978 to 2011 for training, data from 2012 to 2018 for validation, and data from 2019 to 2024 for testing. 

Our models use $N_{in}=12$ monthly averaged Antarctic SIC fields to 
autoregressively predict up to $N_{out}=12$ future months. To quantitatively evaluate the performance of the proposed structure, 
a binary land mask is applied to restrict the quantification in forecasting domain $A_{\Omega}$.
Root mean squared error (RMSE) is calculated directly from the continuous SIC 
regression outputs. The predicted SIC fields are further divided into three 
classes: open water ($\mathrm{SIC}<15\%$), marginal ice 
($15\%\leq\mathrm{SIC}\leq80\%$), and full ice ($\mathrm{SIC}>80\%$) \cite{icener-ori,batte2020summer,zampieri2018bright}, 
from which three-class accuracy (Cls3 Acc), Macro-F1 score (mF1) , and F1 score are computed. We additionally 
report binary accuracy (Cls2 Acc) based on the 15\% ice threshold:
\begin{equation*}
    \text{Binary Accuracy}
    =
    \left(1-\frac{\mathrm{IIEE}}{A_{\Omega}}\right)\times100\%,
\end{equation*}
where IIEE is the integrated ice-edge error \cite{goessling2016predictability}.  

\section{Methods: Convolutional Transformer Structure}
\label{sec:fram}

\begin{figure}[t]
    \centering
    \includegraphics[width=\linewidth]{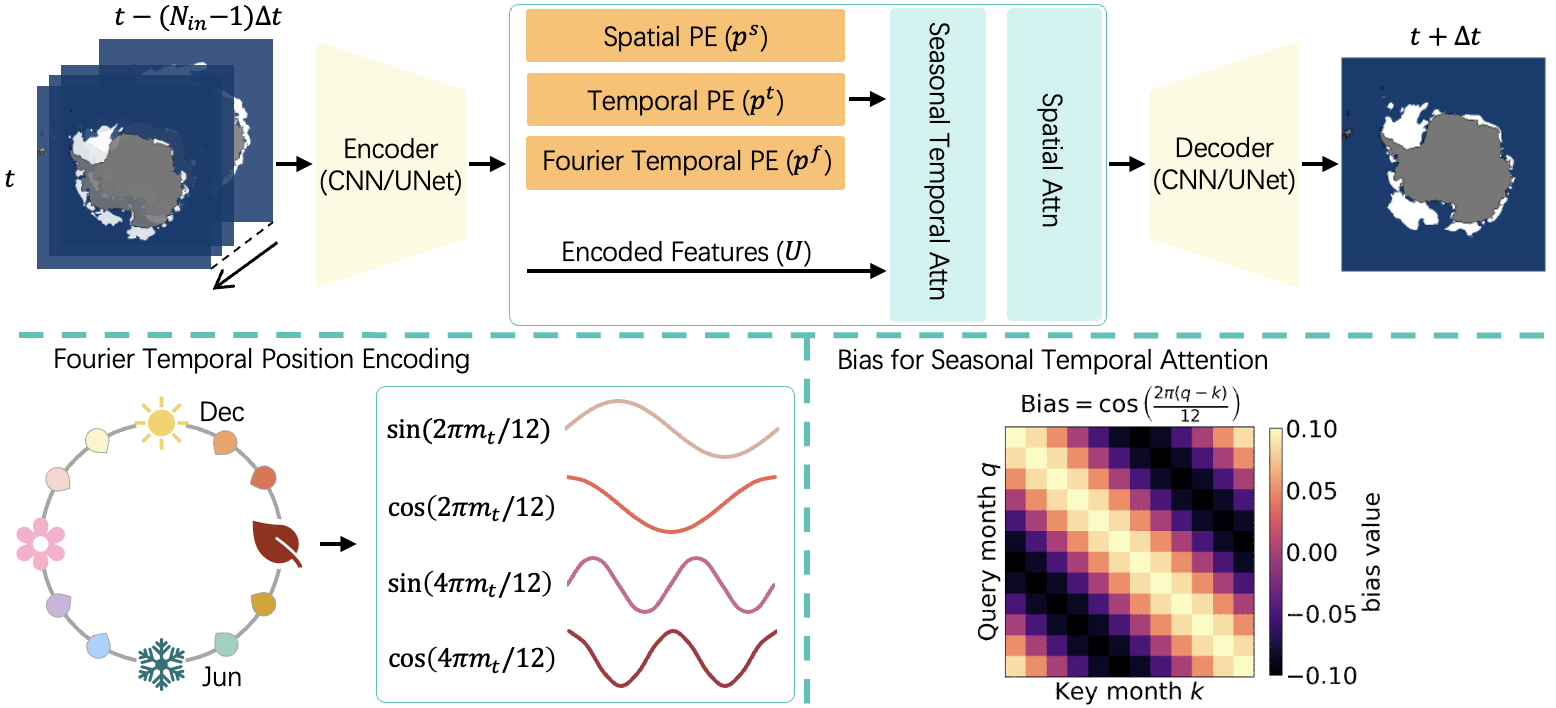}
    \caption{Overview of the proposed one-step SIC forecasting model and detailed components. Historical SIC fields are encoded into features, augmented with spatial and temporal positional encodings, processed by seasonal temporal and spatial attention, and decoded to predict SIC at \(t+\Delta t\). The left lower panel shows the Fourier temporal encoding, and the right lower panel shows the seasonal bias constructed from the circular distance between query and key months.}
    \label{fig:structure}
\end{figure}

To address the spatial blurring and accumulating errors in long-horizon SIC forecasting, we propose a spatio-temporal architecture that combines convolutional feature extraction with a factorised self-attention bottleneck.  The convolutional encoder-decoder learns local spatial representations of SIC fields, while the temporal and spatial attention modules capture long-range dependencies and seasonally varying evolution patterns. 
The model produce continuous SIC forecasts for autoregressive rollout and discrete sea-ice state predictions for classification-based evaluation. We refer to Figure \ref{fig:structure} for the overall structure. In the following subsections, we follow the top row of Figure \ref{fig:structure} from left to right and outline the detailed design of our proposed structure.

\subsection{Tokenisation of SIC Fields}
Let $\mathbb{X} := \{ \mathcal{X} \}\in \mathbb{R}^{N_{in} \times H\times W}$ denote a batch of historical SIC sequences. We first design a frame-wise CNN encoder for the SIC monthly field (i.e. in $H \times W$) denoted as $E(\cdot): \mathbb{R}^{ H\times W} \rightarrow\mathbb{R}^{C' \times H'\times W'}$, with a downsampling depth $d$ to extract local spatial features and obtain a compact bottleneck representation.
The encoder $E(\cdot)$ is applied to monthly SIC fields to obtain the feature space as:
\begin{equation*}
    \{E(X_{t})\in\mathbb{R}^{C' \times H'\times W'}, \quad \forall X_t \in \mathbb{X} \}, 
\end{equation*}
where $C'$ is the bottleneck channel dimension, $H', W'$ are bottleneck spatial resolution. 
We flatten the spatial dimensions of $E(X_t)$ into $S=H'W'$ tokens with the set 
\begin{equation*}
U:=\{\mathrm{reshape}(E(X_{t}))\}_{X_t \in \mathbb{X}} \in\mathbb{R}^{N_{in} \times C' \times S}.
\end{equation*}
This tokenisation reduces the full spatio-temporal token length from $N_{in}HW$ at the image resolution to $N_{in}S$ at the bottleneck resolution, where $S \ll HW$. The resulting token sequence enables efficient factorised spatio-temporal attention, while the convolutional encoder-decoder retains local spatial structure for fine-scale sea-ice prediction.

\subsection{Month-Aware Positional Encoding}

Spatio-temporal Transformer models commonly incorporate spatial and temporal positional encodings to distinguish tokens at different spatial locations and time steps \cite{bertasius2021space, arnab2021vivit}. However, SIC forecasting exhibits strong seasonal periodicity: same calendar months across different years share similar large-scale seasonal dynamic behaviour. Thus, in addition to spatial and temporal positional encodings, we introduce month-aware positional encoding. The position encoded token can be written as 
\begin{equation} \label{encoder}
    Z_{t,n} = U_{t,n}+\mathbf{p}^{s}_{n}+\mathbf{p}^{t}_{t}+\mathbf{p}^{f}_{m_t},
\qquad Z_{t,n}\in\mathbb{R}^{C'},
\end{equation}
where $t\in\{0,1,\cdots,11\}$ and $n\in\{1,2,\cdots,S\}$ are indices of input timestep and spatial token, respectively; $m_f\in\{0,\ldots,11\}$ denotes the calendar month associated with timestep $t$.
 
Rather than relying on a single positional representation, \eqref{encoder} separates physical priors into three distinct embeddings: (i) Let $\mathbf{P}^{s}\in\mathbb{R}^{S\times C'}$ represent the full spatial embedding table, and $\mathbf{p}^{s}_{n}=\mathbf{P}^{s}(n)\in\mathbb{R}^{C'}$ is the embedding for spatial token $n$. The spatial positional encoding is implemented as a learnable tensor which is added to the bottleneck features and shared across all time steps. (ii) $\mathbf{P}^{t}\in\mathbb{R}^{N_{in}\times C'}$ denotes the fixed sinusoidal temporal positional encoding, and $\mathbf{p}^{t}_{t}=\mathbf{P}^{t}(t)\in\mathbb{R}^{C'}$. (iii) $\mathbf{p}^{f}_{m_t}\in\mathbb{R}^{C'}$ denotes the Fourier-based calendar-month embedding, referred to MonthPE in the following. For each month index $m_{t} = {0, \dots,11}$, we define a Fourier basis:
\begin{equation*}
\Phi(m_t)= \left[ \sin\left(\frac{2\pi m_t}{12}\right), \cos\left(\frac{2\pi m_t}{12}\right), \dots,
\sin\left(\frac{2\pi N_f m_t}{12}\right),\, \cos\left(\frac{2\pi N_f m_t}{12}\right) \right]^{\top}
\in \mathbb{R}^{2N_f},
\end{equation*}
where $N_f$ denotes the number of Fourier modes.
In our implementation, we set $N_f=4$, which allows the month encoding to retain the annual, semiannual, and several higher-order seasonal harmonics. This choice provides a balance between expressiveness and simplicity: it is sufficiently rich to represent non-sinusoidal seasonal patterns in monthly sea-ice evolution, while avoiding unnecessarily high-frequency components for a 12-month calendar.
The Fourier basis is then mapped into the token space by a learnable projection $W^f$,
\begin{equation*}
\mathbf{p}^{f}_{m_{t}}=W^f \Phi(m_t),\qquad
W^f\in\mathbb{R}^{C'\times 2N_f}.
\end{equation*}
The resulting MonthPE is added to every spatial token at time step $t$ as in \eqref{encoder}, allowing the model to use an explicit 12-month periodic prior knowledge while learning how to mix different seasonal harmonics.

\subsection{Factorised Spatio-Temporal Attention}
Computing global self-attention over the full spatio-temporal token sequence of length $N_{in}S$ incurs an attention complexity of $\mathcal{O}((N_{in}S)^2 C')$, which becomes prohibitive for high-resolution inputs (when $S \gg 1$) and long temporal contexts (when $N_{in} \gg 1$). Following \cite{bertasius2021space}, we therefore adopt a divided space-time attention scheme, where attention is factorised into a temporal operator and a spatial operator applied sequentially. Specifically, temporal attention is performed independently for each spatial token, and spatial attention is then performed independently for each time step. This mechanism reduces the attention complexity to $\mathcal{O}(N_{in}^2SC' + S^2N_{in}C')$, where the first term corresponds to temporal attention and the second to spatial attention.

For notational convenience, we denote the temporal and spatial multi-head attention layers by $\mathcal{A}^{t}(\cdot)$ and $\mathcal{A}^{s}(\cdot)$, respectively, where each operator includes the corresponding attention computation and output projection. The two attention operations are applied sequentially with residual connections:
\begin{equation*}
Z^{t} = Z + \mathcal{A}^{t}(Z), \quad Z' = Z^{t} + \mathcal{A}^{s}(Z^{t}),
\end{equation*}
where $Z^{t}$ denotes the temporally representation before spatial attention layers and $Z'$ is the output of the divided space-time attention block. Both attention operators preserve the token embedding dimension $C'$.

For spatial attention, each timestep is processed independently across the $S$ spatial tokens. For each attention head, $Q^{s},K^{s},V^{s} \in\mathbb{R}^{S\times d_h}$, where $d_h$ denotes the dimension of each attention head. Standard scaled dot-product attention is applied as 
\begin{equation*}
    \text{Attn}^{s}(Q^{s},K^{s},V^{s})=\text{Softmax}\left(\frac{Q^{s}K^{s\top}}{\sqrt{d_h}}\right)V^{s}.
\end{equation*}

The SIC latent representation $Z^s$ from $Z$ in \eqref{encoder} is used in the query, key and value matrices as:
\begin{equation*}
    Q^s= Z^sW_Q, \quad K^s=Z^sW_K, \quad V^s=Z^sW_V,
\end{equation*}
where \(W_Q\), \(W_K\), and \(W_V\) are learnable projection matrices.

Temporal attention is applied independently for each spatial token $n$. For each batch element, layer, and attention head,  $Q^{t}, K^{t}, V^{t}\in \mathbb{R}^{N_{in}\times d_h}$.
Unlike using the standard attention structure in spatial representation, 
to explicitly incorporate annual periodicity into temporal interactions, 
we introduce a seasonal bias into the temporal attention logits. 

Let $q,k\in \{0,\dots,11\}$ denote the index associated with query timestep $q$ and key timestep $k$. We define a pairwise bias matrix $b = (b_{qk})$ as 
\begin{equation} \label{eq:bqk}
b_{qk} = \alpha \cos\left(\frac{2\pi (q-k)}{12}\right), \qquad \alpha\in\mathbb{R}, 
\end{equation}
where $\alpha$ is a learnable scalar controlling the strength of the seasonal attention bias.
In particular, $\alpha$ is learned separately for each temporal attention layer. 
The magnitude of the bias $b$ implies attention strength.
A positive value indicates the attention of two nearby months, where a negative value indicates the attention of two season-opposite months. For each batch element, spatial token and attention head, a bias $b$ is learned as in \eqref{eq:bqk}.
The attention is then computed as
\begin{equation*}
\text{Attn}^t(Q^t,K^t,V^t) = \text{Softmax}\left(\frac{Q^tK^{t\top}}{\sqrt{d_h}} + b\right)V^t,
\end{equation*}

Unlike positional encodings, which are injected into token representations, the proposed seasonal bias directly modulates the attention scores, thereby shaping the temporal dependency structure at the level of query-key interactions. The cosine form provides a simple and smooth periodic prior over the annual cycle, making it a natural choice for monthly geophysical data with strong seasonality. 
Since $\alpha$ is learned jointly with the rest of the network, the model can adaptively determine how strongly this seasonality-awareness should influence attention.

\subsection{Decoder and Autoregressive Forecasting}
To project the spatio-temporal tokens back into the physical domain, we apply a convolutional decoder, denoted as $D(\cdot)$. Following the attention bottleneck, the latent sequence is reshaped from $N_{in}SC'$ back to the spatial grid $\mathbb{R}^{N_{in} \times C' \times H' \times W'}$.
The decoder maps the latent representation to the next-step SIC prediction,
\begin{equation*}
\hat{X}_{t+\Delta t}=D(Z').
\end{equation*}
We consider both CNN-based and UNet-based backbone variants. For the CNN-based variant, the decoder reconstructs the output directly from the bottleneck representation without multi-scale skip connections. For the UNet-based variant, encoder features at corresponding spatial resolutions are passed to the decoder through skip connections, providing additional high-resolution spatial information during reconstruction.

Together with the encoder and spatio-temporal attention blocks described above, the complete one-step forecasting model is denoted by $\mathcal{T}_{\theta}$.
Given the preceding $N_{in}$ monthly SIC fields, the one-step prediction is written as
\begin{equation*}
    \hat{X}_{t+\Delta t}= \mathcal{T}_{\theta}\left(\{X_{t-(N_{in}-1)\Delta t},\dots,X_t\} \right).  
\end{equation*}

To achieve long-horizon polar forecasting, we employ an autoregressive step. 
For forecast month $\ell=1,\ldots,N_{out}$, the model is rolled out autoregressively as
\begin{equation} \label{eq:Ttheta}
    \hat{X}_{t+\ell\Delta t} =\mathcal{T}_{\theta}\left(\{\tilde{X}_{t-(N_{in}-\ell)\Delta t},\dots,\tilde{X}_{t+(\ell-1)\Delta t}\}\right), 
\end{equation}
where
\begin{equation*}
\tilde{X}_{\tau}=\begin{cases}X_{\tau}, & \tau \leq t,\\
                \hat{X}_{\tau}, & \tau > t.
                \end{cases}, \qquad \tau \in [t-(N_{in}-1)\Delta t,t+(N_{out}-1)\Delta t].
\end{equation*} 
The overall forecasting operator $\mathcal{F}_\theta(\cdot): \mathbb{R}^{ N_{in}\times H\times W} \rightarrow\mathbb{R}^{N_{out}\times H\times W}$ is then defined as \eqref{eq:Ttheta} with $\ell=1,\ldots,N_{out}$.

Lastly, to train our neural network, we apply a loss function defined as
\begin{align*}
    L^{\text{total}} & = w\big(L^{\text{mse}}(\hat{X}_{t+\Delta t},X_{t+\Delta t}) + L^{\text{mse}}(\hat{X}_{t+2\Delta t},X_{t+2\Delta t})\big) \nonumber\\
    & \quad + (1-w)\big(L^{\text{foc}}(\hat{X}_{t+\Delta t},X_{t+\Delta t}) + L^{\text{foc}}(\hat{X}_{t+2\Delta t},X_{t+2\Delta t})\big),
\end{align*}
where $L^{\text{mse}}$ is the standard mean squared error (MSE), $L^{\text{foc}}$ is the Focal loss for classification with parameters to be specified in Section \ref{sec:exp}, and $w \in (0,1)$ is the weight of the loss components. $L^{\text{mse}}(\hat{X}_{t+2\Delta t},X_{t+2\Delta t})$ and $L^{\text{foc}}(\hat{X}_{t+2\Delta t},X_{t+2\Delta t})$ are designed to mitigate the accumulating errors due to the autoregressive step in \eqref{eq:Ttheta}. 
Moreover, this improves the robustness and stability of the autoregression for longer time forecasting.

\section{Experiments}
\label{sec:exp}

\begin{figure}[ht]
    \centering
    \includegraphics[width=\linewidth]{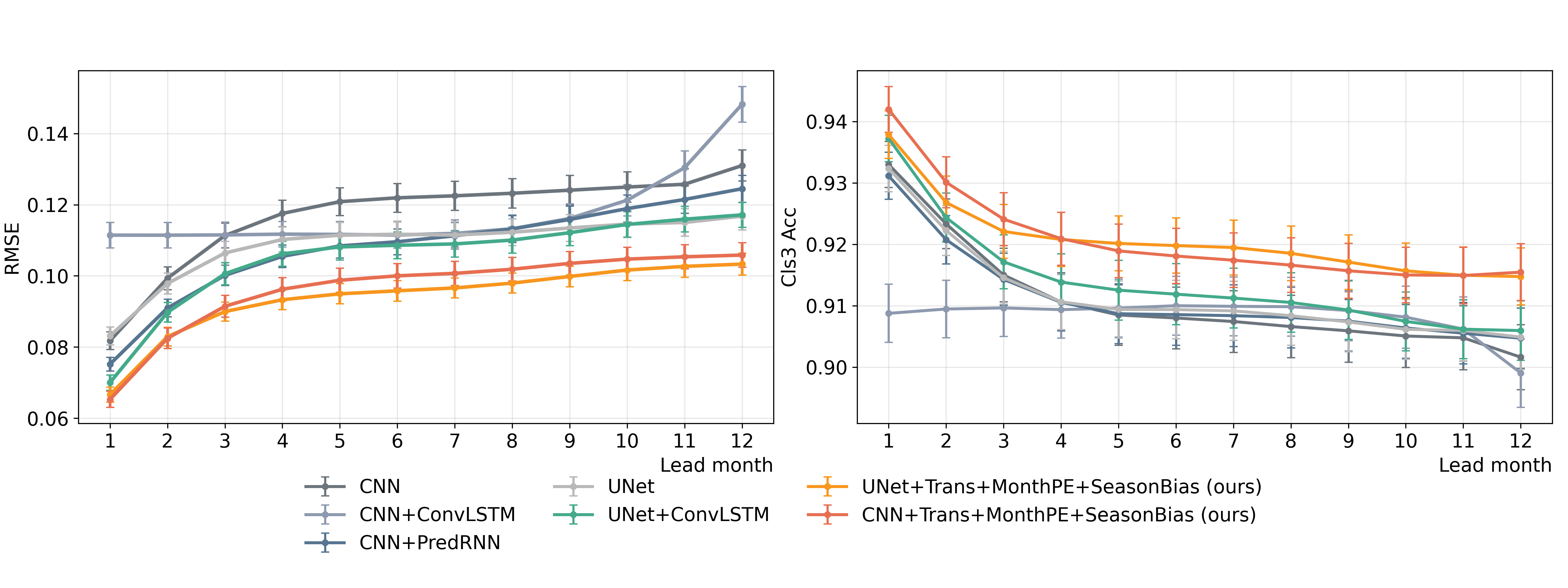}
    \caption{Comparison of multi-horizon forecasting performance across baseline and Transformer-based models over a 12-month autoregressive prediction. The left panel shows the RMSE for continuous SIC prediction, while the right panel shows the classification accuracy for categorical SIC prediction. Markers denote the mean over test samples, and error bars represent the standard error.}
\label{fig:multi_horizon_results}
\end{figure}
We evaluate the proposed hybrid framework through ablation studies and comparisons with several baseline models, including CNN, UNet, ConvLSTM, PredRNN and Transformer-based variants \cite{ronneberger2015u,shi2015convolutional,wang2022predrnn,bertasius2021space}. The experimental setup and key hyperparameters are summarised in Table \ref{tab:hyperparameters}, other settings and detailed structures are includes in Appendix \ref{App_1}. For training, the classification branch uses Focal Loss to address class imbalance, with class weights set to $[0.2, 0.3, 0.5]$ for open water, marginal ice, and full ice, respectively. The regression branch is optimised using MSE loss. The two losses are combined with weights of $0.1$ and $0.9$ for the classification and regression objectives. All experiments are conducted on machines with single NVIDIA V100 GPUs.

\subsection{Overall Performance Comparison}

We first compare the proposed model with representative baselines to evaluate its overall forecasting performance. Figure \ref{fig:multi_horizon_results} shows the performance of different deep learning models over a 12-month autoregressive forecasting horizon. As the forecast lead time increases, RMSE generally increases while classification accuracy decreases, reflecting uncertainty accumulation and error propagation during autoregressive rollout. Nevertheless, both proposed seasonality-aware Transformer-based models with UNet and CNN backbones achieve the lowest or near-lowest RMSE and the highest or near-highest classification accuracy throughout the forecasting horizon, consistently outperforming the remaining baselines across the forecasting horizon. These results suggest that combining convolutional spatial encoding with Transformer-based temporal modelling and explicit seasonal prior information is beneficial for both continuous and categorical Antarctic SIC forecasting. A qualitative comparison of the 2023 SIC classification forecasts is provided in Appendix \ref{App_2}.

In addition, we consider the comparison with three simple statistical methods: the same calendar month of the previous year (Last-year), monthly climatology, and anomaly persistence \cite{10.1093/oso/9780199202782.003.0015, https://doi.org/10.1029/2021JC017784}. We further compare against SEAS5 over its available six-month forecast horizon and a multi-source UNet baseline inspired by IceNet and ANTSIC-UNet \cite{icener-ori,tc-19-6381-2025}. The latter uses 12 months of historical SIC, recent climate variables, and 6 months of linear-trend SIC predictions. Except for Last-year, all learning-based baselines use the training split described in Section \ref{sec:prob}; all methods, including SEAS5, are evaluated over the same test period. As shown in Table \ref{tab:baseline_comparison}, even without additional climate-variable inputs, the proposed models consistently outperform the statistical baselines and SEAS5 across the evaluated forecast horizons, while achieving comparable or better performance than the ANTSIC-UNet-inspired baseline. These results demonstrate the effectiveness of the proposed framework for both short- and long-range forecasting.

\begin{table*}[t]
\centering
\caption{Comparison with statistical, dynamical, and multi-source baselines at lead months 1, 6, and 12.}
\label{tab:baseline_comparison}
\resizebox{\textwidth}{!}{
\begin{tabular}{lcccccc}
\toprule
Method & RMSE@1 $\downarrow$ & RMSE@6 $\downarrow$ & RMSE@12 $\downarrow$ & Cls3 Acc@1 $\uparrow$ & Cls3 Acc@6 $\uparrow$ & Cls3 Acc@12 $\uparrow$ \\
\midrule
Last-year & 0.1248 & 0.1255 & 0.1292 & 0.9066 & 0.9087 & 0.9061 \\
Monthly climatology & 0.1056 & 0.1080 & 0.1080 & 0.9113 & 0.9111 & 0.9107 \\
Anomaly persistence & 0.0876 & 0.1261 & 0.1292 & 0.9261 & 0.8833 & 0.9061 \\
SEAS5 & 0.0734 & 0.1279 & - & 0.9368 & 0.8972 & - \\
ANTSIC-UNet-inspired & 0.0719 & 0.0983 & - & 0.9382 & \textbf{0.9215} & - \\
\midrule
CNN+Trans+MonthPE+SeasonBias & \textbf{0.0653} & 0.1000 & 0.1058 & \textbf{0.9420} & 0.9181 & \textbf{0.9155} \\
UNet+Trans+MonthPE+SeasonBias & 0.0666 & \textbf{0.0958} & \textbf{0.1033} & 0.9379 & 0.9198 & 0.9147 \\
\bottomrule
\end{tabular}
}
\end{table*}

\subsection{Ablation study at Short- and Long-Horizon Forecast Steps}

\begin{table*}[ht]
\centering
\caption{Performance of different methods when the lead times of one and twelve months.}
\label{tab:test_step_results}
\resizebox{\textwidth}{!}{
\begin{tabular}{llccccc}
\toprule
Backbone & Add on & Cls3 Acc & Cls3 mF1 & Cls2 Acc & Cls2 F1 & RMSE \\
\midrule
\multicolumn{7}{l}{\textbf{Lead 1}} \\
\midrule
CNN+Trans &
- & 0.9371$\pm$0.0039 & 0.8166$\pm$0.0060 & 0.9727$\pm$0.0012 & 0.9148$\pm$0.0081 & 0.0679$\pm$0.0020 \\
 & 
+MonthPE 
& 0.9392$\pm$0.0041 & 0.8216$\pm$0.0062 & 0.9745$\pm$0.0012 & 0.9209$\pm$0.0071 & 0.0658$\pm$0.0022 \\
&
+SeasonBias 
& 0.9378$\pm$0.0040 & 0.8182$\pm$0.0062 & 0.9738$\pm$0.0011 & 0.9173$\pm$0.0078 & 0.0672$\pm$0.0022 \\
&
+MonthPE+SeasonBias 
& \textbf{0.9420}$\pm$0.0037 & \textbf{0.8280}$\pm$0.0063 & \textbf{0.9751}$\pm$0.0012 & \textbf{0.9234}$\pm$0.0069 & \textbf{0.0653}$\pm$0.0022 \\
\midrule
UNet+Trans &
-
& 0.9353$\pm$0.0036 & 0.8087$\pm$0.0066 & 0.9703$\pm$0.0013 & 0.9046$\pm$0.0094 & 0.0714$\pm$0.0021 \\
&
+MonthPE 
& 0.9256$\pm$0.0044 & 0.7827$\pm$0.0086 & 0.9682$\pm$0.0017 & 0.9011$\pm$0.0109 & 0.0724$\pm$0.0024 \\
&
+SeasonBias 
& 0.9340$\pm$0.0038 & 0.8035$\pm$0.0067 & 0.9712$\pm$0.0013 & 0.9073$\pm$0.0092 & 0.0710$\pm$0.0021 \\
&
+MonthPE+SeasonBias 
& \textbf{0.9379}$\pm$0.0039 & \textbf{0.8119}$\pm$0.0067 & \textbf{0.9739}$\pm$0.0013 & \textbf{0.9190}$\pm$0.0073 & \textbf{0.0666}$\pm$0.0021 \\
\midrule
\multicolumn{7}{l}{\textbf{Lead 12}} \\
\midrule
CNN+Trans &
- & 0.9036$\pm$0.0047 & 0.6995$\pm$0.0123 & 0.9486$\pm$0.0020 & 0.8159$\pm$0.0192 & 0.1169$\pm$0.0035 \\
 & 
+MonthPE 
& 0.9067$\pm$0.0052 & 0.7239$\pm$0.0089 & 0.9522$\pm$0.0019 & 0.8395$\pm$0.0151 & 0.1125$\pm$0.0034 \\
&
+SeasonBias 
& 0.9098$\pm$0.0046 & 0.7119$\pm$0.0112 & 0.9511$\pm$0.0019 & 0.8063$\pm$0.0206 & 0.1194$\pm$0.0046 \\
&
+MonthPE+SeasonBias 
& \textbf{0.9155}$\pm$0.0046 & \textbf{0.7490}$\pm$0.0086 & \textbf{0.9561}$\pm$0.0019 & \textbf{0.8583}$\pm$0.0129 & \textbf{0.1058}$\pm$0.0034 \\
\midrule
UNet+Trans & -
& 0.9080$\pm$0.0049 & 0.7223$\pm$0.0101 & 0.9531$\pm$0.0020 & 0.8486$\pm$0.0135 & 0.1114$\pm$0.0038 \\
 & 
+MonthPE 
& 0.9123$\pm$0.0043 & \textbf{0.7510}$\pm$0.0077 & 0.9551$\pm$0.0015 & 0.8571$\pm$0.0141 & \textbf{0.1026}$\pm$0.0028 \\
&
+SeasonBias 
& 0.9089$\pm$0.0048 & 0.7288$\pm$0.0096 & 0.9538$\pm$0.0018 & 0.8543$\pm$0.0128 & 0.1063$\pm$0.0036 \\
&
+MonthPE+SeasonBias 
& \textbf{0.9147}$\pm$0.0047 & 0.7476$\pm$0.0077 & \textbf{0.9576}$\pm$0.0017 & \textbf{0.8615}$\pm$0.0127 & 0.1033$\pm$0.0031 \\
\toprule
\end{tabular}
}
\end{table*}

To study the effect of MonthPE and SeasonBias, we compare model performance at both the first (lead time one month) and the last (lead time twelve months) autoregressive prediction steps, as reported in Table \ref{tab:test_step_results}.
This comparison allows us to assess the contribution of the seasonality-aware components in both short-range prediction and long-horizon autoregressive forecasting.

When forecasting one month lead time, the combined use of MonthPE and SeasonBias achieves the best performance for both CNN+Trans and UNet+Trans across all reported metrics. For CNN+Trans, MonthPE alone improves all metrics over the baseline, while SeasonBias provides moderate gains; their combination further improves the results, reducing the regression RMSE from 0.0679 to 0.0653 and increasing the Cls3 Acc from 0.9371 to 0.9420. A similar trend is observed for UNet+Trans, where the combined setting achieves the best classification and regression performance, reducing RMSE from 0.0714 to 0.0666. These results indicate that the two seasonality components provide complementary short-range benefits by injecting calendar-month information and encouraging attention to seasonally related historical states.

When forecasting 12 months SIC fields, the benefit of explicit seasonality-awareness becomes more evident under long-horizon autoregressive rollout.
For CNN+Trans, the combination of MonthPE and SeasonBias consistently achieves the best performance, improving Cls3 Acc from 0.9036 to 0.9155 and reducing RMSE from 0.1169 to 0.1058, indicating that the proposed components help stabilise long-range prediction. For UNet+Trans, the combined setting achieves the best Cls3 Acc, Cls2 Acc, and Cls2 F1, while MonthPE alone gives slightly better Cls3 mF1 and RMSE. Therefore, unlike the CNN-based backbone, the UNet-based backbone shows a more metric-dependent response at the final rollout step. This may be because the UNet skip connection already provides strong multi-scale spatial representations. Nevertheless, both MonthPE and SeasonBias generally improve long-horizon performance over the baseline, demonstrating the effectiveness of seasonal prior information for monthly Antarctic SIC forecasting. Additional ablation studies on SeasonBias strength and input sequence length are provided in Appendix \ref{App_3}.

\subsection{Temporal Attention}
\label{attn}
\begin{figure}[ht]
    \centering
    \includegraphics[width=\linewidth]{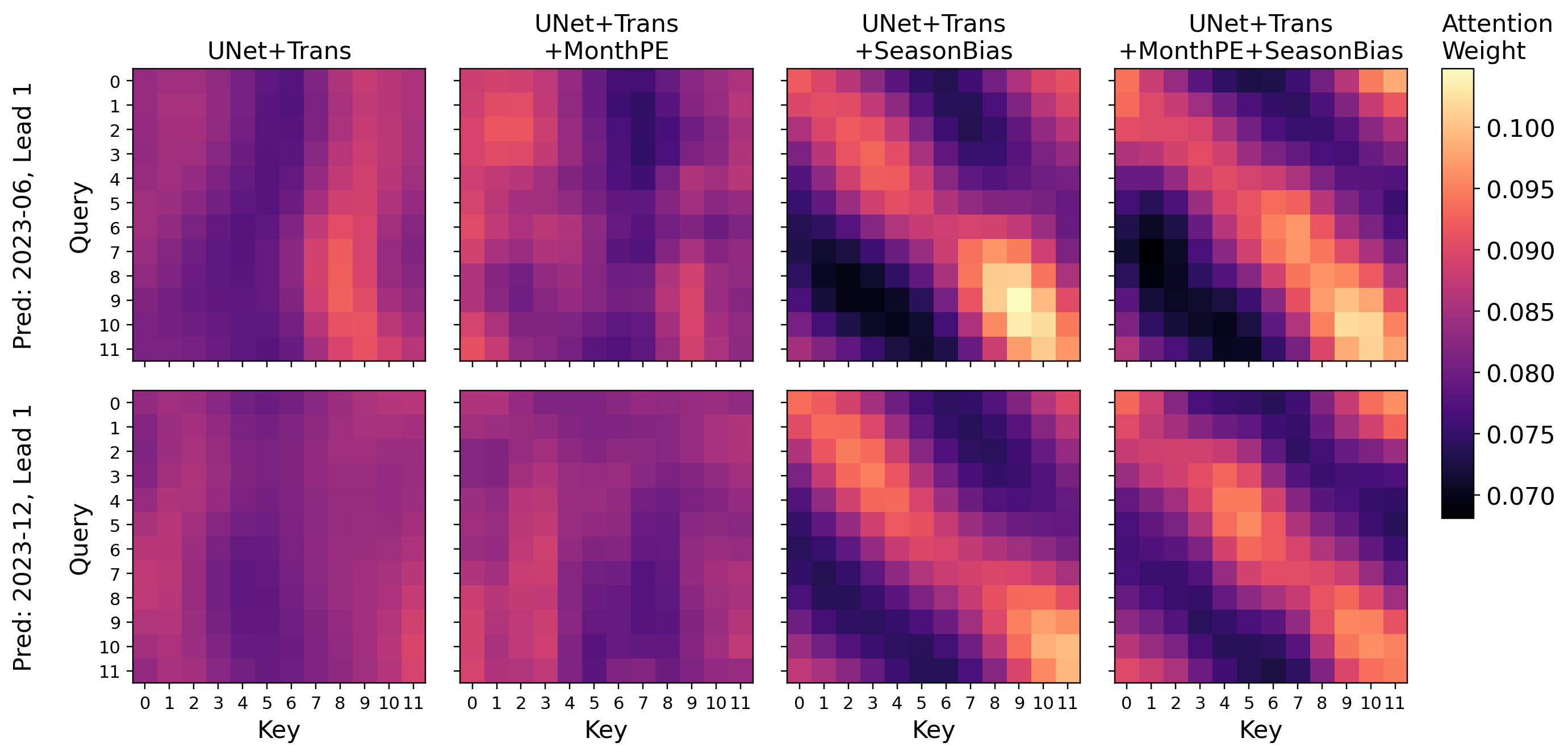}
    \caption{Temporal attention heatmaps of UNet+Trans variants for representative forecast cases. Columns correspond to the baseline model, the model with MonthPE, the model with SeasonBias, and the model with both MonthPE and SeasonBias. Rows show predictions for June and December 2023 with one month lead time.}
    \label{fig:temporal_attention}
\end{figure}

To better understand how MonthPE and SeasonBias affect the temporal behaviour of the Transformer bottleneck, we visualise the temporal attention correlations of different UNet+Trans variants in Figure \ref{fig:temporal_attention}.
The baseline UNet+Trans model produces relatively diffuse attention patterns, with only weak preference for particular key months. Adding MonthPE introduces clearer month-dependent variation, suggesting that explicit calendar-month information affects how the model organises temporal memory, although the attention remains comparatively smooth. In contrast, the models with SeasonBias show much clearer diagonal and banded structures. This indicates that the temporal attention is no longer distributed uniformly over the input sequence, but is guided towards seasonally related query-key pairs. The combined MonthPE+SeasonBias model preserves this periodic structure while also allowing the attention weights to vary with the target month. This behaviour is consistent with the design of SeasonBias, which directly modulates the temporal attention logits according to the circular distance between calendar months.

These visualisations provide qualitative evidence that the proposed seasonality-aware components affect the internal temporal routing of the Transformer bottleneck. MonthPE injects seasonality information into the token representation, whereas SeasonBias more directly reshapes the query-key attention pattern. Together, they encourage more seasonally structured temporal dependencies, which is consistent with the improved forecasting performance observed in the quantitative results.

\section{Conclusions}
\label{sec:con}
In this paper, we presented a hybrid convolutional transformer framework for Antarctic SIC forecasting. The proposed model combines convolutional spatial representation learning with Transformer-based spatio-temporal dependency modelling, and further incorporates seasonal information through Fourier-based positional encoding and seasonal temporal bias attention.
Experimental results showed that the architecture achieves competitive forecasting performance across both classification and regression metrics. The ablation study further indicated that Fourier-based positional encoding and seasonal bias provides consistent benefits in both short- and long-range prediction. Overall, these results demonstrate the potential of hybrid convolutional and attention-based models for Antarctic SIC forecasting, especially when seasonal information is incorporated appropriately.

Our work has certain limitations. The current framework is primarily data-driven and does not explicitly use physical knowledge of the Earth system, such as thermodynamic constraints, sea-ice dynamics, or coupled ocean-atmosphere information. Prior studies suggest that incorporating physical constraints or scientific knowledge into learning-based models can improve physical consistency and robustness \cite{karniadakis2021physics, willard2022integrating, icener-ori}. Such information may help improve the prediction of fine-scale regions, including sharp ice-edge structures and local heterogeneous patterns. In addition, since the experiments are conducted on historical Antarctic SIC data, the robustness of the model under rare extreme events and future climate regimes remains uncertain. Future work will investigate physics-guided extensions and evaluate the proposed design across broader geophysical prediction settings.

For future work, the proposed design may also be applicable to a broader range of Earth science prediction problems with strong seasonal or cyclic behaviour, and more generally to time-series forecasting tasks with known periodic patterns. Another direction is to develop more systematic interpretability analyses for attention-based SIC forecasting models. While the attention correlations in this work provide a qualitative view of the learned temporal and spatial dependencies, future studies could compare these patterns across seasons, regions, and forecast horizons to assess whether the model consistently relies on physically meaningful information. Such analyses may help identify when the model captures stable seasonal correlations.

% \bibdata{ref}
% \bibstyle{plain}
\bibliography{ref}
\bibliographystyle{plain}

\appendix
\section{Network Structure and Parameters}
\label{App_1}
Additional implementation details of the network architectures are provided as follows. The main text focuses on the proposed seasonality-aware components, while this appendix summarises the corresponding backbone structures, bottleneck designs, and parameter counts for reproducibility. All models use the same input-output setting and are evaluated under the same training and testing protocol unless otherwise stated.

Table \ref{tab:arch_detail_unet} and  \ref{tab:arch_detail_cnn} summarise the detailed structure of the proposed convolutional transformer model. The convolutional encoder first extracts local spatial features and reduces the spatial resolution before the Transformer bottleneck is applied. At the bottleneck, the features are converted into spatio-temporal tokens, where positional encodings and factorised self-attention are used to model temporal and spatial dependencies. The decoder then upsamples the features back to the original spatial resolution and produces both regression and classification outputs.

Table \ref{tab:hyperparameters} summarises the main forecasting, model, seasonal-prior, training, and loss settings. All compared models use the same input resolution, historical input window, forecast horizon, loss functions, and training protocol unless otherwise specified.
Table \ref{tab:bottleneck_details} further compares the bottleneck modules used in different baselines. The pure CNN and UNet baselines rely on convolutional bottlenecks without a temporal modelling module. The ConvLSTM and PredRNN variants introduce recurrent temporal modelling at the bottleneck feature level. The Transformer-based variants reshape the bottleneck features into spatio-temporal tokens and apply factorised temporal and spatial self-attention. 
The main architectural difference among all models lies in the bottleneck module, while other settings and training strategy are kept consistent. Since the attention modules are applied only after spatial downsampling at the bottleneck resolution, all experiments can be run on a single NVIDIA V100 GPU with gradient checkpointing, without requiring model parallelism.

\begin{table}
\centering
\small
\caption{Detailed architecture of the proposed UNet-Transformer model with MonthPE and SeasonBias. 
Here $N_{in}=12$, $H=W=352$, $S=44\times44$, and $C_{in}=256$. The batch dimension is omitted for readability.}
\label{tab:arch_detail_unet}
\resizebox{\textwidth}{!}{
\begin{tabular}{l|l|c|c}
\toprule
\textbf{Block} & \textbf{Layer} & \textbf{Resolution} & \textbf{Channels} \\
\midrule

Input 
& -- 
& $352 \times 352$ 
& $1$ \\
\midrule

\multirow{4}{*}{Encoder Stem}
& Conv $3\times3$
& $352 \times 352$ 
& $1 \rightarrow 32$ \\ 
& ReLU 
& $352 \times 352$ 
& $32$ \\
& Conv $3\times3$
& $352 \times 352$ 
& $32$\\ 
& ReLU 
& $352 \times 352$ 
& $32$ \\
\midrule

\multirow{5}{*}{Encoder Down 1}
& MaxPool $2\times2$ 
& $352 \times 352 \rightarrow 176 \times 176$ 
& $32$ \\
& Conv $3\times3$
& $176 \times 176$ 
& $32 \rightarrow 64$\\
& ReLU 
& $176 \times 176$ 
& $64$ \\
& Conv $3\times3$ 
& $176 \times 176$ 
& $64$\\
& ReLU 
& $176 \times 176$ 
& $64$ \\
\midrule

\multirow{5}{*}{Encoder Down 2}
& MaxPool $2\times2$ 
& $176 \times 176 \rightarrow 88 \times 88$ 
& $64$ \\
& Conv $3\times3$
& $88 \times 88$ 
& $64 \rightarrow 128$\\
& ReLU 
& $88 \times 88$ 
& $128$ \\
& Conv $3\times3$
& $88 \times 88$ 
& $128$ \\
& ReLU 
& $88 \times 88$ 
& $128$ \\
\midrule

\multirow{5}{*}{Encoder Down 3}
& MaxPool $2\times2$ 
& $88 \times 88 \rightarrow 44 \times 44$ 
& $128$ \\
& Conv $3\times3$
& $44 \times 44$ 
& $128 \rightarrow 256$ \\
& ReLU 
& $44 \times 44$ 
& $256$ \\
& Conv $3\times3$
& $44 \times 44$ 
& $256$ \\
& ReLU 
& $44 \times 44$ 
& $256$ \\
\midrule

\multirow{4}{*}{Bottleneck Embedding}
& Learnable spatial PE 
& $44 \times 44$ 
& $256$ \\
& Flatten to tokens 
& $44 \times 44 \rightarrow S=1936$ 
& $256$ \\
& Relative time PE 
& $N_{in} \times S$ 
& $256$ \\
& MonthPE 
& $N_{in} \times S$ 
& $256$ \\
\midrule

\multirow{10}{*}{Transformer Block $\times 2$}
& Temporal reshape 
& $N_{in} \times S \rightarrow S \times N_{in}$ 
& $256$ \\
& LayerNorm 
& $S \times N_{in}$ 
& $256$ \\
& Temporal MHA 
& $S \times N_{in}$ 
& $256$ \\
& Residual 
& $S \times N_{in}$ 
& $256$ \\
& Spatial reshape 
& $S \times N_{in} \rightarrow N_{in} \times S$ 
& $256$ \\
& LayerNorm 
& $N_{in} \times S$ 
& $256$ \\
& Spatial MHA 
& $N_{in} \times S$ 
& $256$ \\
& Residual 
& $N_{in} \times S$ 
& $256$ \\
& FFN
& $N_{in} \times S$ 
& $256 \rightarrow 1024 \rightarrow 256$ \\
& Residual 
& $N_{in} \times S$ 
& $256$ \\
\midrule

\multirow{6}{*}{Decoder Up 1}
& ConvTranspose $2\times2$ 
& $44 \times 44 \rightarrow 88 \times 88$ 
& $256 \rightarrow 128$ \\
& Concat skip 
& $88 \times 88$ 
& $128 + 128 = 256$ \\
& Conv $3\times3$
& $88 \times 88$ 
& $256 \rightarrow 128$ \\
& ReLU
& $88 \times 88$ 
& $128$ \\
& Conv $3\times3$
& $88 \times 88$ 
& $128$ \\
& ReLU 
& $88 \times 88$ 
& $128$ \\
\midrule

\multirow{6}{*}{Decoder Up 2}
& ConvTranspose $2\times2$ 
& $88 \times 88 \rightarrow 176 \times 176$ 
& $128 \rightarrow 64$ \\
& Concat skip 
& $176 \times 176$ 
& $64 + 64 = 128$ \\
& Conv $3\times3$
& $176 \times 176$ 
& $128 \rightarrow 64$ \\
& ReLU
& $176 \times 176$ 
& $64$ \\
& Conv $3\times3$
& $176 \times 176$ 
& $64$ \\
& ReLU 
& $176 \times 176$ 
& $64$ \\
\midrule

\multirow{6}{*}{Decoder Up 3}
& ConvTranspose $2\times2$ 
& $176 \times 176 \rightarrow 352 \times 352$ 
& $64 \rightarrow 32$ \\
& Concat skip 
& $352 \times 352$ 
& $32 + 32 = 64$ \\
& Conv $3\times3$
& $352 \times 352$ 
& $64 \rightarrow 32$ \\
& ReLU
& $352 \times 352$ 
& $32$ \\
& Conv $3\times3$
& $352 \times 352$ 
& $32$ \\
& ReLU 
& $352 \times 352$ 
& $32$ \\
\midrule

\multirow{4}{*}{Output Heads}
& Classification head ($1\times1$ Conv) 
& $352 \times 352$ 
& $32 \rightarrow C_{\mathrm{cls}}$ \\
& Regression head ($1\times1$ Conv) 
& $352 \times 352$ 
& $32 \rightarrow 1$ \\
& Classification temporal weighting 
& $352 \times 352$ 
& $C_{\mathrm{cls}}$ \\
& Regression temporal weighting + Sigmoid 
& $352 \times 352$ 
& $1$ \\

\bottomrule
\end{tabular}
}
\end{table}

\begin{table}
\centering
\small
\caption{Detailed architecture of the proposed CNN-Transformer model with MonthPE and SeasonBias. 
Here $N_{in}=12$, $H=W=352$, $S=44\times44$, and $C_{in}=256$. The batch dimension is omitted for readability.}
\label{tab:arch_detail_cnn}
\resizebox{\textwidth}{!}{
\begin{tabular}{l|l|c|c}
\toprule
\textbf{Block} & \textbf{Layer} & \textbf{Resolution} & \textbf{Channels} \\
\midrule

Input 
& -- 
& $352 \times 352$ 
& $1$ \\
\midrule

\multirow{4}{*}{Encoder Stem}
& Conv $3\times3$
& $352 \times 352$ 
& $1 \rightarrow 32$ \\ 
& ReLU 
& $352 \times 352$ 
& $32$ \\
& Conv $3\times3$
& $352 \times 352$ 
& $32$\\ 
& ReLU 
& $352 \times 352$ 
& $32$ \\
\midrule

\multirow{5}{*}{Encoder Down 1}
& MaxPool $2\times2$ 
& $352 \times 352 \rightarrow 176 \times 176$ 
& $32$ \\
& Conv $3\times3$
& $176 \times 176$ 
& $32 \rightarrow 64$\\
& ReLU 
& $176 \times 176$ 
& $64$ \\
& Conv $3\times3$ 
& $176 \times 176$ 
& $64$\\
& ReLU 
& $176 \times 176$ 
& $64$ \\
\midrule

\multirow{5}{*}{Encoder Down 2}
& MaxPool $2\times2$ 
& $176 \times 176 \rightarrow 88 \times 88$ 
& $64$ \\
& Conv $3\times3$
& $88 \times 88$ 
& $64 \rightarrow 128$\\
& ReLU 
& $88 \times 88$ 
& $128$ \\
& Conv $3\times3$
& $88 \times 88$ 
& $128$ \\
& ReLU 
& $88 \times 88$ 
& $128$ \\
\midrule

\multirow{5}{*}{Encoder Down 3}
& MaxPool $2\times2$ 
& $88 \times 88 \rightarrow 44 \times 44$ 
& $128$ \\
& Conv $3\times3$
& $44 \times 44$ 
& $128 \rightarrow 256$ \\
& ReLU 
& $44 \times 44$ 
& $256$ \\
& Conv $3\times3$
& $44 \times 44$ 
& $256$ \\
& ReLU 
& $44 \times 44$ 
& $256$ \\
\midrule

\multirow{4}{*}{Bottleneck Embedding}
& Learnable spatial PE 
& $44 \times 44$ 
& $256$ \\
& Flatten to tokens 
& $44 \times 44 \rightarrow S=1936$ 
& $256$ \\
& Relative time PE 
& $N_{in} \times S$ 
& $256$ \\
& MonthPE 
& $N_{in} \times S$ 
& $256$ \\
\midrule

\multirow{10}{*}{Transformer Block $\times 2$}
& Temporal reshape 
& $N_{in} \times S \rightarrow S \times N_{in}$ 
& $256$ \\
& LayerNorm 
& $S \times N_{in}$ 
& $256$ \\
& Temporal MHA + SeasonBias 
& $S \times N_{in}$ 
& $256$ \\
& Residual 
& $S \times N_{in}$ 
& $256$ \\
& Spatial reshape 
& $S \times N_{in} \rightarrow N_{in} \times S$ 
& $256$ \\
& LayerNorm 
& $N_{in} \times S$ 
& $256$ \\
& Spatial MHA 
& $N_{in} \times S$ 
& $256$ \\
& Residual 
& $N_{in} \times S$ 
& $256$ \\
& FFN
& $N_{in} \times S$ 
& $256 \rightarrow 1024 \rightarrow 256$ \\
& Residual 
& $N_{in} \times S$ 
& $256$ \\
\midrule

\multirow{5}{*}{CNN Decoder Up 1}
& ConvTranspose $2\times2$ 
& $44 \times 44 \rightarrow 88 \times 88$ 
& $256 \rightarrow 128$ \\
& Conv $3\times3$
& $88 \times 88$ 
& $128 \rightarrow 128$ \\
& ReLU
& $88 \times 88$ 
& $128$ \\
& Conv $3\times3$
& $88 \times 88$ 
& $128$ \\
& ReLU 
& $88 \times 88$ 
& $128$ \\
\midrule

\multirow{5}{*}{CNN Decoder Up 2}
& ConvTranspose $2\times2$ 
& $88 \times 88 \rightarrow 176 \times 176$ 
& $128 \rightarrow 64$ \\
& Conv $3\times3$
& $176 \times 176$ 
& $64 \rightarrow 64$ \\
& ReLU
& $176 \times 176$ 
& $64$ \\
& Conv $3\times3$
& $176 \times 176$ 
& $64$ \\
& ReLU 
& $176 \times 176$ 
& $64$ \\
\midrule

\multirow{5}{*}{CNN Decoder Up 3}
& ConvTranspose $2\times2$ 
& $176 \times 176 \rightarrow 352 \times 352$ 
& $64 \rightarrow 32$ \\
& Conv $3\times3$
& $352 \times 352$ 
& $32 \rightarrow 32$ \\
& ReLU
& $352 \times 352$ 
& $32$ \\
& Conv $3\times3$
& $352 \times 352$ 
& $32$ \\
& ReLU 
& $352 \times 352$ 
& $32$ \\
\midrule

\multirow{4}{*}{Output Heads}
& Classification head ($1\times1$ Conv) 
& $352 \times 352$ 
& $32 \rightarrow C_{\mathrm{cls}}$ \\
& Regression head ($1\times1$ Conv) 
& $352 \times 352$ 
& $32 \rightarrow 1$ \\
& Classification temporal weighting 
& $352 \times 352$ 
& $C_{\mathrm{cls}}$ \\
& Regression temporal weighting + Sigmoid 
& $352 \times 352$ 
& $1$ \\

\bottomrule
\end{tabular}
}
\end{table}

\newpage
\begin{table}[ht]
\centering
\caption{Hyperparameters and experimental setup.}
\label{tab:hyperparameters}
\begin{tabular}{@{}llc@{}}
\toprule
\textbf{Category} & \textbf{Hyperparameter} & \textbf{Value} \\ 
\midrule

\multirow{4}{*}{\textbf{Forecasting}} 
& Input resolution & $352 \times 352$ \\
& Historical input window ($N_{in}$) & 12 months \\
& Forecast horizon ($K$) & 12 months \\
& Output classes & 3(Cls) and 1(Reg) \\ 
\midrule

\multirow{4}{*}{\textbf{Model}} 
& Spatial depth ($d$) & 3 \\
& Temperal Transformer layers & 2 \\
& Spacial Transformer layers & 2 \\
& Attention heads ($h$) & 8 \\
& MLP ratio & 4 \\ 
\midrule

\multirow{3}{*}{\textbf{Seasonality-aware components}} 
& MonthPE Fourier modes ($N_f$) & 4 \\
& SeasonBias initialization ($\alpha_0$) & 0.1 \\
& Attention dropout & 0.0 \\ 
\midrule

\multirow{4}{*}{\textbf{Training}} 
& Optimiser & Adam \\
& Learning rate & $1\times10^{-5}$ \\
& Batch size & 1 \\
& Training epochs & 200 \\ 
\midrule

\multirow{3}{*}{\textbf{Loss}} 
& Classification loss & Focal loss \\
& Regression loss & MSE loss \\
& Loss weights [Cls, Reg] & [0.1, 0.9] \\ 
\bottomrule
\end{tabular}
\end{table}

\begin{table}[ht]
\centering
\caption{Implementation details of the bottleneck modules used in the compared models. 
Here $N_{in}=12$, $S=44\times44=1936$, and $C=256$. 
'-' indicates that no explicit temporal bottleneck module is used.}
\label{tab:bottleneck_details}
\resizebox{\textwidth}{!}{
\begin{tabular}{l|c|c|c}
\toprule
\textbf{Model} 
& \textbf{Bottleneck input} 
& \textbf{Bottleneck module} 
& \textbf{Implementation details} \\
\midrule

CNN 
& $N_{in}\times C\times44\times44$
& -
& standard convolutional bottleneck \\
\midrule
UNet 
& $N_{in}\times C\times44\times44$
& -
& standard UNet bottleneck with skip connections \\
\midrule
CNN+ConvLSTM 
& $N_{in}\times C\times44\times44$
& ConvLSTM
& \begin{tabular}{c}
ConvLSTM down/up bottleneck \\
hidden channels $=256$ \\
kernel size $=3\times3$
\end{tabular} \\
\midrule
UNet+ConvLSTM 
& $N_{in}\times C\times44\times44$
& ConvLSTM
& \begin{tabular}{c}
ConvLSTM down/up bottleneck \\
hidden channels $=256$ \\
kernel size $=3\times3$
\end{tabular} \\
\midrule
CNN+PredRNN 
& $N_{in}\times C\times44\times44$
& PredRNN
& \begin{tabular}{c}
PredRNN bottleneck \\
hidden channels $=256$\\
\end{tabular} \\
\midrule
CNN+Trans+MonthPE+SeasonBias 
& $N_{in}\times S\times C$
& Factorised Transformer
& \begin{tabular}{c}
Position Encoder\\
temporal MHA, heads $=8$ \\
spatial MHA, heads $=8$ \\
\end{tabular} \\
\midrule
UNet+Trans+MonthPE+SeasonBias 
& $N_{in}\times S\times C$
& Factorised Transformer
& \begin{tabular}{c}
Position Encoder\\
temporal MHA, heads $=8$ \\
spatial MHA, heads $=8$ \\
\end{tabular} \\

\bottomrule
\end{tabular}
}
\end{table}

\clearpage
\newpage

\section{Qualitative Comparison}
\label{App_2}

\begin{figure}[ht]
    \centering
    \includegraphics[width=\linewidth]{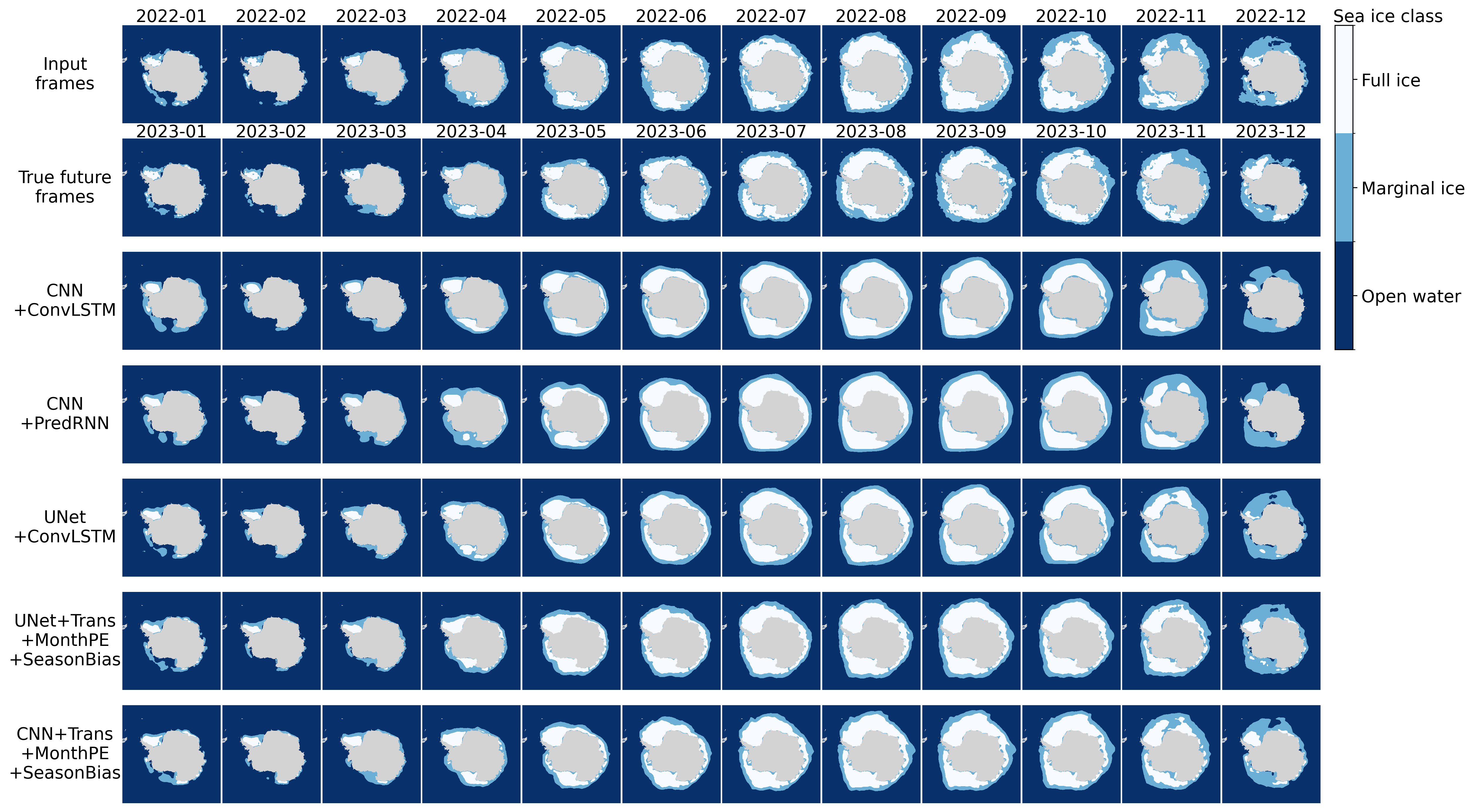}
    \caption{Qualitative comparison of SIC classification forecasts for 2023. The first row shows the 12-month input sequence from 2022, and the second row shows the corresponding ground-truth SIC classes for 2023. The remaining rows compare the predicted SIC classes generated by different forecasting models. Grey regions indicate masked land.}
    \label{fig:qualitative_prediction}
\end{figure}

To further examine the spatial characteristics of the forecasts, Figure \ref{fig:qualitative_prediction} provides a qualitative comparison of SIC classification for the 2023 prediction period. All models capture the dominant seasonal expansion and retreat of Antarctic sea ice, but visible differences appear near the marginal ice zone, where classification is more difficult due to stronger spatial variability.
The recurrent and convolutional baselines tend to produce smoother or less accurate ice-edge structures, whereas the Transformer-based models with MonthPE and SeasonBias better preserve the seasonal evolution of the ice boundary.
This visual comparison is consistent with the quantitative results in Section \ref{sec:exp},Figure \ref{fig:multi_horizon_results}, showing that explicit seasonality-awareness improve the model's ability to represent both the temporal and spatial structure of SIC evolution.

\section{Additional ablation studies}
\label{App_3}

We perform additional ablation studies on SeasonBias strength.
Table \ref{tab:bias_step12_results} reports the twelve lead months forecasting performance of the UNet+Trans backbone with MonthPE under different initial SeasonBias strengths. The results show that the model is relatively robust to the choice of $\alpha$, with only moderate variations across the tested range. Among the tested settings, $\alpha=0.12$ achieves the best overall performance, obtaining the highest Cls3 Acc, Cls3 mF1, Cls2 Acc, Cls2 F1, and the lowest regression RMSE. This suggests that a moderate seasonal bias provides the most effective prior for long-range forecasting, while a too weak or too strong bias may slightly reduce either classification or regression performance. 

Table \ref{tab:step1_step12_results} evaluates the effect of increasing the input window from 12 to 24 months. Although the model maintains reasonable forecasting performance at both one lead month and twelve lead months, the longer input window does not provide a clear overall improvement over the 12-month input setting. This suggests that simply extending the historical context is not necessarily beneficial for SIC forecasting, and that the model may require more effective mechanisms to select or weight useful temporal information from longer sequences.

\begin{table}
\centering
\caption{Twelve lead months forecasting performance under different SeasonBias strengths (mean $\pm$ SE over 48 test samples) on UNet+Trans backbone with MonthPE.}
\label{tab:bias_step12_results}
\resizebox{\textwidth}{!}{%
\begin{tabular}{lccccc}
\toprule
SeasonBias & Cls3 Acc & Cls3 mF1 & Cls2 Acc & Cls2 F1 & RMSE \\
\midrule
$\alpha=0.04$
& 0.9068$\pm$0.0050 
& 0.7178$\pm$0.0090 
& 0.9518$\pm$0.0018 
& 0.8420$\pm$0.0136 
& 0.1160$\pm$0.0038 \\

$\alpha=0.08$
& 0.9066$\pm$0.0052 
& 0.7091$\pm$0.0104 
& 0.9536$\pm$0.0020 
& 0.8494$\pm$0.0136 
& 0.1093$\pm$0.0033 \\

$\alpha=0.12$
& \textbf{0.9117}$\pm$0.0047 
& \textbf{0.7402}$\pm$0.0082 
& \textbf{0.9538}$\pm$0.0020 
& \textbf{0.8538}$\pm$0.0127 
& \textbf{0.1075}$\pm$0.0032 \\

$\alpha=0.16$
& 0.9093$\pm$0.0048 
& 0.7133$\pm$0.0102 
& 0.9527$\pm$0.0019 
& 0.8365$\pm$0.0154 
& 0.1184$\pm$0.0039 \\

$\alpha=0.20$
& 0.9098$\pm$0.0048 
& 0.7082$\pm$0.0122 
& 0.9542$\pm$0.0020 
& 0.8514$\pm$0.0135 
& 0.1108$\pm$0.0032 \\
\bottomrule
\end{tabular}%
}
\end{table}

\begin{table}
\centering
\caption{Forecast performance at one lead month and twelve lead months using 24 input months.}
\label{tab:step1_step12_results}
\resizebox{\textwidth}{!}{%
\begin{tabular}{cccccc}
\toprule
Lead & Cls3 Acc & Cls3 mF1 & Cls2 Acc & Cls2 F1 & RMSE \\
\midrule
1  & $0.9355\pm0.0094$ & $0.8080\pm0.0128$ & $0.9726\pm0.0032$ & $0.9131\pm0.0167$ & $0.0661\pm0.0046$ \\
12 & $0.9146\pm0.0105$ & $0.7527\pm0.0188$ & $0.9569\pm0.0043$ & $0.8691\pm0.0244$ & $0.0988\pm0.0075$ \\
\bottomrule
\end{tabular}
}
\end{table}

% \end{document}

% \newpage
% \subsection*{Additional experiments results}
% \label{App_3}
% Figure \ref{fig:cls_compare_202306} and \ref{fig:cls_compare_202312} provides a qualitative comparison of the multi-step classification forecasts initialized in June and December 2023. All method reproduce the major large-scale sea ice distribution. The visual differences are most apparent near the ice edge, where small spatial shifts or excessive smoothing can change the predicted class.

% \begin{figure}[h]
%     \centering
%     \includegraphics[width=\linewidth]{image/cls_compare_202306_steps1to12.png}
%     \caption{Qualitative comparison of multi-step sea ice classification forecasts initialized in June 2023. 
%     Each column represents a forecast lead time from step 1 to step 12, and each row corresponds to the ground truth or one forecasting method. The three classes denote open water, marginal ice, and full ice. }
%     \label{fig:cls_compare_202306}
% \end{figure}

% \begin{figure}
%     \centering
%     \includegraphics[width=\linewidth]{image/cls_compare_202312_steps1to12.png}
%     \caption{Qualitative comparison of multi-step sea ice classification forecasts initialized in December 2023. 
%     Each column represents a forecast lead time from step 1 to step 12, and each row corresponds to the ground truth or one forecasting method. The three classes denote open water, marginal ice, and full ice. }
%     \label{fig:cls_compare_202312}
% \end{figure}

% \bibdata{ref}
% \bibstyle{plain}

\end{document}